\documentclass[conference]{IEEEtran}
\IEEEoverridecommandlockouts
\usepackage{cite}
\usepackage{amsmath,amssymb,amsfonts}
\usepackage{algorithmic}
\usepackage{graphicx}
\usepackage{textcomp}
\usepackage{xcolor}
\usepackage{titlesec}
\usepackage{cleveref}
\usepackage{url}
\def\BibTeX{{\rm B\kern-.05em{\sc i\kern-.025em b}\kern-.08em
    T\kern-.1667em\lower.7ex\hbox{E}\kern-.125emX}}
\begin{document}

\title{Pseudo-Random UAV Test Generation Using Low-Fidelity Path Simulator\\
\thanks{This work was supported by the EPSRC under Grant EP/V050966/1.}
}

\author{\IEEEauthorblockN{Anas Shrinah}
\IEEEauthorblockA{\textit{School of Computer Science} \\
\textit{University of Bristol}\\
Bristol, UK \\
Anas.Shrinah@bristol.ac.uk}
\and
\IEEEauthorblockN{Kerstin Eder}
\IEEEauthorblockA{\textit{School of Computer Science} \\
\textit{University of Bristol}\\
Bristol, UK \\
Kerstin.Eder@bristol.ac.uk}
}

\maketitle

\begin{abstract}

Simulation-based testing provides a safe and cost-effective environment for verifying the safety of Uncrewed Aerial Vehicles (UAVs).
However, simulation can be resource-consuming, especially when High-Fidelity Simulators (HFS) are used.
To optimise simulation resources, we propose a pseudo-random test generator that uses a Low-Fidelity Simulator (LFS) to estimate UAV flight paths.
Our pseudo-random test generator is the overall winner of the UAV testing competitions in SBFT and ICST 2025 \footnote{\label{note1}\url{https://github.com/skhatiri/UAV-Testing-Competition}}.
This work simplifies the PX4 autopilot HFS to develop a LFS, which operates one order of magnitude faster than the HFS.
Test cases predicted to cause safety violations in the LFS are subsequently validated using the HFS.

\end{abstract}


\section{Introduction}

This paper presents PRGenUAV-LFS, a test generation tool that participated in the UAV Testing Competition organised by the Search-Based and Fuzz Testing (SBFT 2025) workshop \cite{SBFT-UAV2025}.
This competition uses the Aerialist test bench to automate UAV software testing and simulation \cite{icse2024Aerialist}.
Participating test generators should create challenging scenarios by placing obstacles within a flying arena to force the UAV to fly dangerously close to them. A flight path is considered dangerous if the distance between the UAV and any obstacle is less than 1.5m at any time during the flight.
The software under test is the PX4-Avoidance system, an open-source, vision-based autonomous obstacle avoidance system built on top of PX4-Autopilot.
PX4 employs high-fidelity simulation platforms such as Gazebo to perform flight testing.
High-Fidelity Simulators (HFS) are computationally expensive and require significant time to complete simulations.
For instance, simulating a single UAV test case with Aerialist can take several minutes.
To overcome this challenge, we propose to use a Low-Fidelity Simulator (LFS) to estimate flight paths.
This LFS is faster than the PX4 HFS by one order of magnitude and estimates flight paths using the same planner used by the HFS.
This planner is a local planner based on the 3D Vector Field Histogram (3DVFH*) algorithm \cite{baumann2018obstacle}.
Our test generator samples test cases from a predefined range to produce valid and interesting tests.
These tests are evaluated using the LFS to predict the minimum distance between the UAV's estimated path and the obstacles. Only tests with acceptable expected minimum distances are subsequently tested using the HFS.

\begin{figure}[h!]
\includegraphics[height=6.0cm]{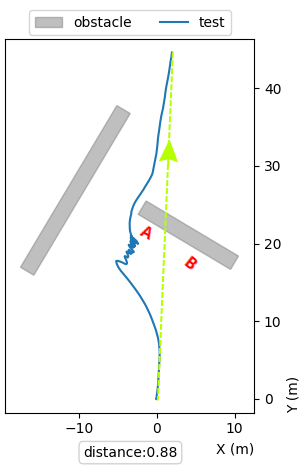}
\centering
\caption{Simulated test case example.}
\label{fig-obstacles-and-soi}
\end{figure}

\section{Pseudo-Random Test Generator}

Test cases are generated for flying missions; each mission consists of a start point, a series of waypoints, and a landing point.
This test generation tool identifies the flight segment closest to the middle line of the Obstacle Designated Arena (ODA) that intersects with the top and bottom borders of the ODA.
After that, the tool places the obstacle to obstruct this segment of interest (SoI).
Obstacle configurations are determined based on the slope and the direction of the SoI.
This section details the process of generating test scenarios for the case shown in~\Cref{fig-obstacles-and-soi}.
Other cases (different slops and directions of the SOI) are handled by applying geometric transformations to the method described below. 
Obstacles are cuboids defined by the coordinates of the centre, length, and width of its base.
We fixed the obstacle height to 20 m, as the UAV is restricted from flying above obstacles.
For simplicity, we refer to an obstacle by its base
We observed that only two obstacles are required to compel the UAV to follow an S-shaped path, which can cause the UAV to fly close to the obstacles.
We chose to set the the width of the bases of the two obstacles to 2m.
The test generator randomly selects the diagonal length, rotation angle, and y-coordinate of the base of the first obstacle from predefined ranges to ensure the the base is fully contained within the designated area.
The base length is then calculated using the diagonal and the width, and the x-coordinate is determined using the y-coordinate and the equation of the line defined by the SoI.
The first obstacle must be placed in the path of the UAV to direct it away from the goal.
This obstacle should intersect the SoI at an obtuse angle and at a point that divides the length of its base into two parts, A and B, as illustrated in~\Cref{fig-obstacles-and-soi}.
The length of part A should be one-third of the base\textquotesingle s length, while part B spans the remaining two-thirds.
This arrangement causes the UAV to track the obstacle along part A, as this path aligns the UAV's heading closer to the goal direction.
This behaviour is enforced by the planner because its cost function favours path nodes with orientations close to the goal direction. 
In \Cref{fig-obstacles-and-soi}, part A (B) is located to the left (right) of the SoI.
To ensure proper placement, one-third (two-thirds) of the diagonal length must be less than the distance between the SoI and the ODA\textquotesingle s left (right) border.
This constraint guarantees that, even if the diagonal is aligned parallel to the X-axis, the vertices of the diagonal remain within the ODA, ensuring the obstacle is horizontally contained.

To ensure vertical containment of the obstacle within the ODA, the y-coordinate must be chosen such that, even if the diagonal is aligned parallel to the Y-axis, the vertices of the diagonal remain inside the ODA.
This is achieved by constraining the y-coordinate plus (minus) half of the diagonal must be less (greater) than the y-coordinate of the ODA\textquotesingle s top (bottom) border.
These constraints consider only the two extreme cases, when the diagonal is parallel to the X or Y axis, because these cases encompass all possible obstacle rotations, ensuring that the entire obstacle remains within the ODA under any orientation.
After flying near the first obstacle, the planner will adjust the UAV\textquotesingle s path to realign its orientation towards the goal (the end of the SoI).
To prevent the UAV from flying away from the first obstacle when correcting its path, the second obstacle is placed perpendicular to the first obstacle and close enough to the first obstacle, but not too close to risk the planner failing to find a feasible solution. 
The x- and y-coordinates of the centre of the second obstacle\textquotesingle s base are specified relative to the first obstacle to achieve this requirement and to the borders of the ODA to avoid having parts of this obstacle outside the ODA.
The length of the second obstacle needs to be long enough to prevent the UAV from circling it from below; it is set to 1.75 of the first obstacle length.
The next section describes the LFS used for evaluating the test cases generated.


\section{low-fidelity path simulator}

The 3DVFH* algorithm uses 3D point clouds from a stereo camera to plan obstacle-avoiding paths. While an HFS generates these point clouds in PX4 simulation, our LFS leverages the Open3D library to produce them \cite{Zhou2018}.
The process begins by creating a scene that contains a floor plane and the two given obstacles.
The scene is then rendered into a depth image relative to the UAV position and orientation.
Finally, a 3D point cloud is extracted from this depth image.
Note that it is crucial to use the same intrinsic and extrinsic camera parameters in the LFS as those used in the HFS to ensure point clouds generated by both systems are similar.
Every time 3DVFH* is invoked, it produces a lookahead tree.
The nodes of this tree form the planned path that guides the UAV towards its goal while avoiding obstacles.
In PX4, the planner operates at a fixed rate of 100 Hz, recalculating the path every 100 ms.
The obstacle avoidance system uses these nodes to generate waypoints, which are issued to the UAV flight controller.
The PX4 simulator then simulates the UAV\textquotesingle s motion under the control of the flight controller and updates the planner with the UAV\textquotesingle s new pose.
On the other hand, the LFS does not operate at a specific rate.
Instead, its planning rate is determined solely by the computation time of the planner, which is way smaller than the  100 ms time between planning cycles in the PX4 simulator.
For example, on a system with Intel i7 CPU, 16 GB RAM, RTX 3070 GPU, the planning time was approximately 8ms.
To maintain behaviour similar to the HFS, the LFS presumes 100 ms have passed after each planning cycle and updates the UAV’s pose accordingly.
The LFS abstracts the flight controller and the dynamics of the UAV and uses the following assumptions to estimate the motion of the UAV.
First, the UAV’s maximum linear velocity is assumed to be 3 m/s, and its maximum yaw velocity is 13.5 degrees per second.
Second, during each LFS simulation step (the presumed 100 ms), the UAV is assumed to travel linearly towards the goal (the first node in the lookahead tree) at the maximum linear velocity if the change in the yaw angle that is required to align the UAV with the goal is less than one-third of its yaw velocity.
Otherwise, the UAV travels a fraction of this maximum distance, proportional to the ratio between the required change in the yaw angle and the yaw velocity. 
These assumptions are based on rough estimations derived from preliminary comparisons of the flight paths produced by the LFS and HFS during the simulation of several test cases. While this simplified dynamic model provides a reasonable approximation, developing a more comprehensive abstract model could significantly enhance the accuracy of flight path estimation. This improvement is reserved for future work.
Our pseudo-random test generator is the overall winner of the UAV testing competitions in SBFT and ICST 2025.
The performance of our UAV test case generator was evaluated during the SBFT 2025 UAV Testing Competition and the results are documented in the competition report published by the organisers and in the competition's GitHub repository \footnotemark[\value{footnote}].

\bibliographystyle{plain} 
\bibliography{ref} 

\end{document}